\documentclass[11pt]{article}
\usepackage{eacl2017}
\usepackage{times}
\usepackage{url}
\usepackage{latexsym}
 \usepackage{graphicx}

\eaclfinalcopy 

\title{Deep Semi-Supervised Learning with Linguistically Motivated \\ Sequence Labeling Task Hierarchies }

\author{Jonathan Godwin \\
  University College London \\
  {\tt ucabjwg@ucl.ac.uk} \\\And
  Pontus Stenetorp \\
  University College London\\
  {\tt p.stenetorp@ucl.ac.uk} \\ \\\And
  Sebastian Riedel \\
  University College London\\
  {\tt s.riedel@ucl.ac.uk} \\}
\date{}

\begin{document}
\maketitle
\begin{abstract}
    In this paper we present a novel Neural Network algorithm for conducting semi-supervised learning for sequence labeling tasks arranged in a linguistically motivated hierarchy.
    This relationship is exploited to regularise the representations of supervised tasks by backpropagating the error of the unsupervised task through the supervised tasks.
    We introduce a neural network where lower layers are supervised by downstream tasks and the final layer task is an auxiliary unsupervised task.
    The architecture shows improvements of up to two percentage points $F_{\beta=1}$ for Chunking compared to a plausible baseline.
\end{abstract}

\section{Introduction}
    It is natural to think of NLP tasks existing in a hierarchy, with each task building upon the previous tasks.
    For example, Part of Speech (POS) is known to be an extremely strong feature for Noun Phrase Chunking, and downstream tasks such as greedy Language Modeling (LM) can make use of information about the syntactic and semantic structure recovered from junior tasks in making predictions.
    
    Conversely, information about downstream tasks should also provide information that aids generalisation for junior downstream tasks, a form of semi-supervised learning.
    Arguably, there is a two-way relationship between each pair of tasks.
    
    Following work such as \newcite{sogaard2016deep}, that exploits such hierarchies in a fully supervised setting, we represent this hierarchical relationship within the structure of a multi-task Recurrent Neural Network (RNN), where junior tasks in the hierarchy are supervised on inner layers and the parameters are jointly optimised during training.
    Joint optimisation within a hierarchical network acts as a form of regularisation in two ways: first, it forces the network to learn general representations within the parameters of the shared hidden layers \cite{ando2005framework}; second, there is a penalty on the supervised junior layers for forming a representation and making predictions that are inconsistent with senior tasks.
    Intuitively, we can see how this can be beneficial - when humans receive new information from one task that is inconsistent with with our internal representation of a junior task we update both representations to maintain a coherent view of the world.
    
\begin{figure}
    \centering
    \includegraphics[scale=0.45]{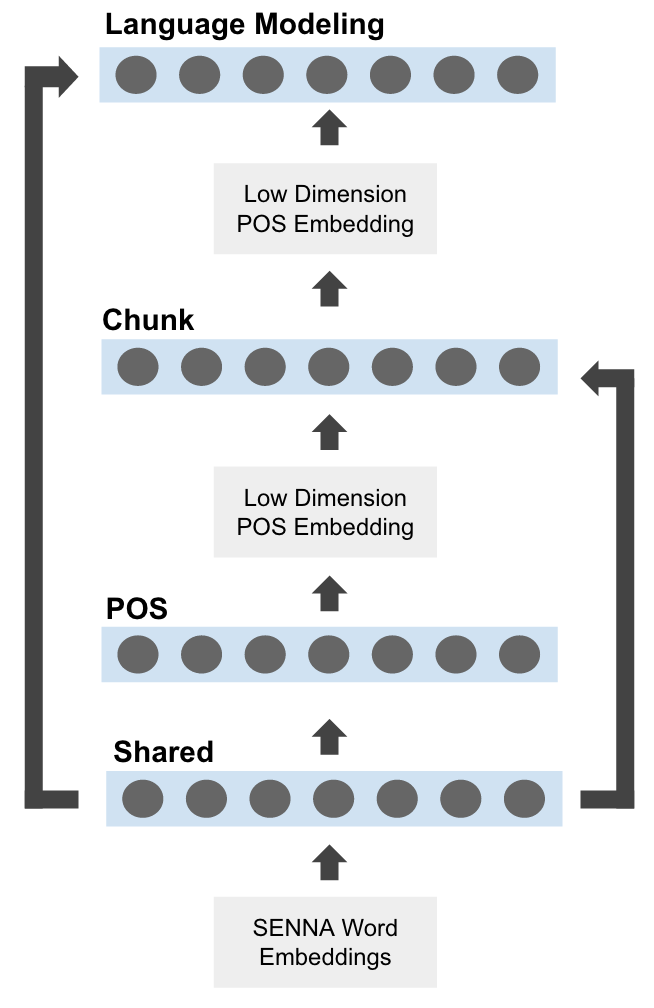}
    \caption{Our Hierarchical Network. In this network, junior tasks are supervised in lower layers, with an unsupervised task (Language Modeling) at the most senior layer.}
    \label{fig:our_model}
\end{figure}

    By incorporating an unsupervised auxiliary task (e.g. \newcite{plank2016multilingual}) as the most senior layer we can use this structure for semi-supervised learning - the error on the unsupervised tasks penalises junior tasks when their representations and predictions are not consistent. It is the aim of this paper to demonstrate that organising a network in such a way can improve performance. To that end, although we do not achieve state of the art results, we see a small but consistent performance improvement against a baseline. A diagram of our model can be seen in Figure \ref{fig:our_model}.
    
    \textbf{Our Contributions:}
    \begin{enumerate}
    \item{we demonstrate a way to use hierarchical neural networks to conduct semi-supervised learning, and demonstrate modest performance improvements against a baseline; and}
    \item{we learn low-dimensional embeddings of labels, and demonstrate that these embeddings capture informative latent structure.}
    \end{enumerate}
    
\section{Linguistically Motivated Task Hierarchies}

When we speak and understand language we are arguably performing many different linguistic tasks at once.
At the top level we might be trying to formulate the best possible sequence of words given all of the contextual and prior information, but this requires us to do lower-level tasks like understanding the syntactic and semantic roles of the words we choose in a specific context. 

This paper seeks to examine the POS tagging, Chunking and Language Modeling hierarchy and demonstrate that, by developing an algorithm that both exploits this structure and optimises all three jointly, we can improve performance.

\subsection{Motivating our Choice of Tasks}

In the original introductory paper to Noun Phrase Chunking, \newcite{abney1991parsing}, Chunking is motivated by describing a three-phase process - first, you read the words and assign a Part of Speech tag, you then use a ‘Chunker’ to group these words together into chunks depending on the context and the Parts of Speech, and finally you build a parse tree on top of the chunks. 

The parallels between this linguistic description of parsing and our architecture are clear; first, we build a prediction for POS, we then use this prediction to assist in parsing by Chunk, which we then use for greedy Language Modeling. In this hierarchy, we consider Language Modeling as auxiliary - designed to improve performance on POS and Chunking, and so therefore results are not presented for this task.

\section{Our Model}

In our model we represent linguistically motivated hierarchies in a multi-task Bi-Directional Recurrent Neural Network where junior tasks in the hierarchy are supervised at lower layers.This architecture builds upon \newcite{sogaard2016deep}, but is adapted in two ways: first, we add an unsupervised sequence labeling task (Language Modeling), second, we add a low-dimensional embedding layer between tasks in the hierarchy to learn dense representations of label tags. In addition to \newcite{sogaard2016deep}. 

Work such as \newcite{mirowski-vlachos:2015:ACL-IJCNLP} in which incorporating syntactic dependencies improves performance, demonstrates the benefits of incorporating junior tasks in prediction.

Our neural network has one hidden layer, after which each successive task is supervised on the next layer.
In addition, we add skip connections from the hidden layer to the senior supervised layers to allow layers to ignore information from junior tasks.

A diagram of our network can be seen in Figure \ref{fig:our_model}.

\subsection{Supervision of Multiple Tasks}

Our model has 3 sources of error signals - one for each task.
Since each task is categorical we use the discrete cross entropy to calculate the loss for each task:

\[
  H(p, q) = - \sum_{i}^{n_{labels}}  p(label_i) \ log \ q(label_i)  
\]

Where $n_{labels}$ is the number of labels in the task, $q(label_i)$ is the probability of label $i$ under the predicted distribution, and $p(label_i)$ is the probability of label $i$ in the true distribution (in this case, a one-hot vector).

During training with fully supervised data (POS, Chunk and Language Modeling), we optimise the mean cross entropy:

\[ 
Loss(x,y) = \frac{1}{n} \sum_{i}^{n} H(y, f_{task_i}(x))
\]

Where $f_{task_i}(x)$ is the predicted distribution on task number $i$ from our model. 

When labels are missing, we drop the associated cross entropy terms from the loss, and omit the cross entropy calculation from the forward pass.

\subsection{Bi-Directional RNNs}

Our network is a Bi-Directional Recurrent Neural Network (Bi-RNN) (\newcite{schuster1997bidirectional}) with Gated Recurrent Units (GRUs) (\newcite{cho2014properties}, \newcite{chung2014empirical}). 

In a Bi-Directional RNN we run left-to-right through the sentence, and then we run right-to-left.
This gives us two hidden states at time step t - one from the left-to-right pass, and one from the right-to-left pass.
These are then combined to provide a probability distribution for the tag token conditioned on all of the other words in the sentence.

\subsection{Implementation Details}

During training we alternate batches of data with POS and Chunk and Language Model labels with batches of just Language Modeling according to some probability $ 0 < \gamma < 1$.

We train our model using the ADAM (\newcite{kingma2014adam}) optimiser for 100 epochs, where one epoch corresponds to one pass through the labelled data.
We train in batch sizes of $32\times32$.

\section{Experimental Results}
\subsection{Data Sets}

We present our experiments on two data sets - CoNLL 2000 Chunking data set (\newcite{tjong2000introduction}) which is derived from the Penn Tree Bank newspaper text (\newcite{marcus1993building}), and the Genia biomedical corpus (\newcite{kim2003genia}), derived from biomedical article abstracts.

These two data sets were chosen since they perform differently under the same classifiers \cite{tsuruoka2005developing}. The unlabelled data for semi-supervised learning for newspaper text is the Penn Tree Bank, and for biomedical text it a custom data set of Pubmed abstracts.

\subsection{Baseline Results}

We compare the results of our model to a baseline multi-task architecture inspired by \newcite{yang2016multi}.
In our baseline model there are no explicit connections between tasks - the only shared parameters are in the hidden layer.

We also present results for our hierarchical model where there is no training on unlabelled data (but there is the LM) and confirm previous results that arranging tasks in a hierarchy improves performance.
Results for both models can be seen for POS in Table \ref{table:hierarchy_summary_results_pos} and for Chunk in Table \ref{table:hierarchy_summary_results_chunk}.

\begin{table}
    \centering
    \resizebox{\columnwidth}{!}{%
        \begin{tabular}{l|r|r|}
        & CoNLL ($F_{\beta=1}$) & Genia ($F_{\beta=1}$)  \\\hline
        \newcite{sogaard2016deep} & 95.6 \% & - \\\hline
        Baseline Model & 92.0 \%  & \textbf{85.5} \% \\ 
        Explicit Task Hierarchy (ETH) & \textbf{92.2} \%  (+ 0.2 \%) & 86.2 \% (+ 0.7 \%) \\ 
        \end{tabular}
        }
    \caption{Hierarchy Chunking Results}
    \label{table:hierarchy_summary_results_chunk}
\end{table}

\begin{table}
    \centering
    \resizebox{\columnwidth}{!}{%
        \begin{tabular}{l|r|r|}
        & CoNLL (Accuracy) & Genia (Accuracy)  \\\hline
        \newcite{tsuruoka2005developing} & - & 98.3 \% \\
        \newcite{ling2015finding} & 97.78 \% & - \\\hline
        Baseline Model & 96.2 \%  & 97.3 \% \\ 
        Explicit Task Hierarchy (ETH) & \textbf{96.4} \% (+ 0.2 \%) & \textbf{97.6} \% (+ 0.3 \%) \\ 
        \end{tabular}
        }
    \caption{Hierarchy POS Results}
    \label{table:hierarchy_summary_results_pos}
\end{table}

\subsection{Semi-Supervised Experiments}

Experiments showing the effects of our semi-supervised learning regime on models initialised both with and without pre-trained word embeddings can be seen in Tables \ref{table:semi_summary_results_chunk} and \ref{table:semi_summary_results_pos}.

In models without pre-trained word embeddings we see a significant improvement associated with the semi-supervised regime.

However, we observe that for models with pre-trained word embeddings, the positive impact of semi-supervised learning is less significant.
This is likely due to the fact some of the regularities learned using the language model are already contained within the embedding.
In fact, the training schedule of SENNA is similar to that of neural language modelling (\newcite{collobert2011natural}). 

Two other points are worthy of mention in the experiments with 100 \% of the training data.
First, the impact of semi-supervised learning on biomedical data is significantly less than on newspaper data.
This is likely due to the smaller overlap between vocabularies in the training set and vocabularies in the test set.
Second, the benefits for POS are smaller than they are for Chunking - this is likely due to the POS weights being more heavily regularised by receiving gradients from both the Chunking and Language Modeling loss.

\begin{table}
    \centering
    \resizebox{\columnwidth}{!}{%
        \begin{tabular}{l|r|r|}
        & CoNLL ($F_{\beta=1}$) &  Genia ($F_{\beta=1}$) \\\hline
        \textbf{No Pre-trained Embeddings} & &\\
        \ Explicit Task Hierarchy (ETH)  & 86.2 \%  & 85.3 \% \\ 
        \ Semi-Supervised \& ETH  & \textbf{88.3} \% (+ 2.1 \%) & \textbf{85.7} \% (+ 0.4 \%) \\
        \textbf{Pre-trained Embeddings} & &\\
        \ Explicit Task Hierarchy (ETH) & 92.0 \% & \textbf{86.2} \% \\
        \ Semi-Supervised \& ETH  & \textbf{92.3} \% (+ 0.4 \%) & 85.9 \% (- 0.3 \%)  \\
        \end{tabular}
        }
    \caption{Chunking Unlabelled Data Results}
    \label{table:semi_summary_results_chunk}
\end{table}

\begin{table}
    \centering
    \resizebox{\columnwidth}{!} {
        \begin{tabular}{l|r|r|}
        & CoNLL (Accuracy) &  Genia (Accuracy) \\ \hline
        \textbf{No Pre-trained Embeddings} & &\\
        \ Explicit Task Hierarchy (ETH) & 92.4 \%  & 97.5 \%  \\ 
        \ Semi-Supervised \& ETH  & \textbf{94.2} \% (+ 1.8 \%) & \textbf{97.6} \% (+ 0.1 \%) \\ 
        \textbf{Pre-trained Embeddings} & & \\
        \ Explicit Task Hierarchy (ETH) & 96.4 \% & 97.3 \% \\
        \ Semi-Supervised \& ETH &  \textbf{96.4} \% (+ 0.0 \%) & \textbf{97.6} \% (+ 0.3 \%) \\
        \end{tabular}
    }
    \caption{POS Unlabelled Data Results}
    \label{table:semi_summary_results_pos}
\end{table}

Finally, we run experiments with only a fraction of the training data to see whether our semi-supervised approach makes our models more robust (Tables \ref{table:semi_summary_results_chunk} and \ref{table:semi_summary_results_pos}).
Here, we find variable but consistent improvement in the performance of our tasks even at 1 \% of the original training data.

\begin{table}
    \centering
    \resizebox{\columnwidth}{!} {
        \begin{tabular}{l|r|r|}
        & CoNLL ($F_{\beta=1}$) &  Genia ($F_{\beta=1}$) \\ \hline
        \textbf{Pre-trained Embeddings} & & \\
        \ \textbf{ETH} &  & \\
        \ \ 100 \% & 92.0 \% & \textbf{86.2} \% \\
        \ \ 75 \% & 92.0 \% & 85.4 \% \\
        \ \ 50 \% & 91.7 \% & 86.1 \% \\
        \ \ 25 \% & 90.5 \% & 83.6 \% \\
        \ \ 1 \% & 74.5 \% & 76.4 \% \\
        \ \textbf{Semi-Supervised} & & \\
        \ \ 100 \% &  \textbf{92.5} \% (+ 0.5 \%) & 85.9 \% (- 0.3 \%) \\
        \ \ 75 \% &  \textbf{92.5} \% (+ 0.5 \%) & \textbf{85.6} \% (+ 0.2 \%) \\
        \ \ 50 \% &  \textbf{91.8} \% (+ 0.1 \%) & \textbf{86.4} \% (+ 0.3 \%) \\
        \ \ 25 \% &  \textbf{90.8} \% (+ 0.3 \%) & \textbf{83.7} \% (+ 0.1 \%) \\
        \ \ 1 \% & \textbf{75.2} \% (+ 0.7 \%) & \textbf{76.6} \% (+ 0.2 \%)\\
        
        \end{tabular}
    }
    \caption{Chunk Semi-Supervised Results}
    \label{table:semi_summary_results_chunk}
\end{table}

\begin{table}
    \centering
    \resizebox{\columnwidth}{!} {
        \begin{tabular}{l|r|r|}
        & CoNLL ($F_{\beta=1}$) &  Genia ($F_{\beta=1}$) \\ \hline
        \textbf{Pre-trained Embeddings} & & \\
        \ \textbf{ETH} &  & \\
        \ \ 100 \% & 96.2 \% &  97.5 \% \\
        \ \ 75 \% & 94.1 \% & \textbf{97.4} \%\\
        \ \ 50 \% & 93.4 \% & 96.7 \% \\
        \ \ 25 \% & 92.1 \% & 95.9 \% \\
        \ \ 1 \% & 75.4 \% & 86.0 \% \\
        \ \textbf{Semi-Supervised} & & \\
        \ \ 100 \% &  \textbf{96.2} \% (+ 0.0 \%) & \textbf{97.6} \% (+ 0.1 \%)\\
        \ \ 75 \% &  \textbf{94.0} \% (- 0.1 \%) & 97.2 \% (- 0.2 \%) \\
        \ \ 50 \% &  \textbf{93.5} \% (+ 0.1 \%) & \textbf{96.8} \% (+ 0.1 \%) \\
        \ \ 25 \% &  \textbf{92.2} \% (+ 0.1 \%) & \textbf{96.4} \% (+ 0.4 \%)\\
        \ \ 1 \% & \textbf{75.2} \% (- 0.2 \%) & \textbf{86.8} \% (+ 0.8 \%) \\
        
        \end{tabular}
    }
    \caption{POS Semi-Supervised Results}
    \label{table:semi_summary_results_pos}
\end{table}

\subsection{Label Embeddings}

Our model structure includes an embedding layer between each task.
This layer allows us to learn low-dimensional vector representations of labels, and expose regularities in a way similar to e.g. \newcite{mikolov2013distributed}.

\begin{figure}
    \centering
    \includegraphics[scale=0.4]{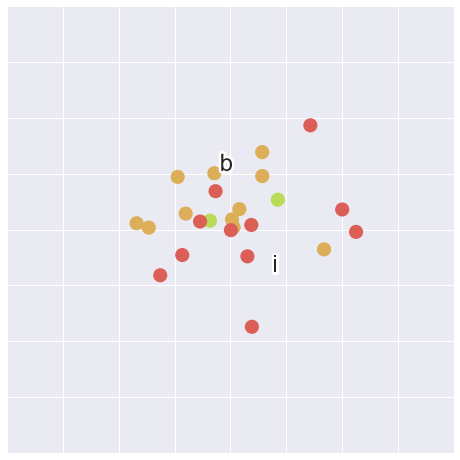}
    \caption{T-SNE for Chunk labels. The orange spots represent labels at the beginning of chunks (`b'), whereas red spots represent labels at the end of chunks (`i'). We can clearly see clusters along the diagonal.  }
    \label{fig:chunk_tsne}
\end{figure}

We demonstrate this in Figure \ref{fig:chunk_tsne} where we present a T-SNE visualisation of our label embeddings for Chunking and observe clusters along the diagonal.

\section{Conclusions \& Further Work}

In this paper we have demonstrated two things: a way to use hierarchical neural networks to conduct semi-supervised learning and the associated performance improvements, and a way to learn low-dimensional embeddings of labels.

Future work would investigate how to address Catastrophic Forgetting \cite{french1999catastrophic} (the problem in Neural Networks of forgetting previous tasks when training on a new task), which leads to the requirement for the mix parameter $\gamma$ in our algorithm, and prevents such models such as ours from scaling to larger supervised task hierarchies where the training data may be various and disjoint.

\bibliography{eacl2017}
\bibliographystyle{eacl2017}

\end{document}